# FedPNN: One-shot Federated Classification via Evolving Clustering Method and Probabilistic Neural Network hybrid

Polaki Durga Prasad[1,2], Yelleti Vivek[1], Vadlamani Ravi[1,*]

[1]*Center for Artificial Intelligence and Machine Learning,*

*Institute For Development And Research In Banking Technology (IDRBT), Hyderabad 500076, India.*

[2]*SCIS, University of Hyderabad, Hyderabad 500046, India.*

21mcmt27@uohyd.ac.in; yvivek@idrbt.ac.in; vravi@idrbt.ac.in

**Abstract**

Protecting data privacy is paramount in the fields such as finance, banking, and healthcare. Federated Learning (FL) has attracted widespread attention due to its decentralized, distributed training and the ability to protect the privacy while obtaining a global shared model. However, FL presents challenges such as communication overhead, and limited resource capability. This motivated us to propose a two-stage federated learning approach toward the objective of privacy protection, which is a first-of-its-kind study as follows: (i) During the first stage, the synthetic dataset is generated by employing two different distributions as noise to the vanilla conditional tabular generative adversarial neural network (CTGAN) resulting in modified CTGAN, and (ii) In the second stage, the Federated Probabilistic Neural Network (FedPNN) is developed and employed for building globally shared classification model. We also employed synthetic dataset metrics to check the quality of the generated synthetic dataset. Further, we proposed a meta-clustering algorithm whereby the cluster centers obtained from the clients are clustered at the server for training the global model. Despite PNN being a one-pass learning classifier, its complexity depends on the training data size. Therefore, we employed a modified evolving clustering method (ECM), another one-pass algorithm to cluster the training data thereby increasing the speed further. Moreover, we conducted sensitivity analysis by varying Dthr, a hyperparameter of ECM at the server and client, one at a time. The effectiveness of our approach is validated on four finance and medical datasets.

# 1. Introduction

As regards the data privacy and security, the advent of a new paradigm of machine learning (ML), known as Federated Learning [1], facilitates training machine learning (ML) models without transferring sensitive data to the server. It supports distributed training in remote siloed data centers as well, where the private data resides. In the FL paradigm a central server trains a better quality learning model with the coordination of a federation of participating clients. In a typical FL setting, it is often assumed that many clients possess a small number of non-independent and identically distributed (non-i.i.d) data, have limited computational power, and restricted communication network capabilities.

---

[*] Corresponding Author

In general, FL is broadly categorized into three different types [31], (i) Horizontal federated learning (HFL), (ii) Vertical federated learning (VFL), and (iii) Federated Transfer learning (FTL). HFL refers to sample (or data points) partitioned federated learning. It is applied in a scenario where there is an overlap in the feature space among the clients but differs in the sample space. On the other hand, VFL is applied in a scenario where the sample space is aligned with different clients but differs in the feature space. FTL is applied in a scenario where a small sample space overlaps but differs in the feature space. FTL allows knowledge transfer across different domains; i.e., a trained model from the source domain used to build an efficient model in a specific target domain.

FL is a domain-agnostic methodology. It has been popularized and heavily used in financial industries such as, to detect fraudulent activities associated with credit or debit cards [2, 3]. The fraudulent transactions have been proven to be one of the major causes of financial losses [2]. In these cases, the Federated fraud detection system [3] based on artificial neural network outperformed the fraud detection system that trained on data belonging to a single bank. Additionally, FL based fraud detection systems provide privacy. However, FL based techniques involve various challenges such as statistical heterogeneity of client data, communication overhead, system heterogeneity, and privacy concerns.

**Table 1:** Notation used in the current study

| Notation | Denotes | Notation | Denotes |
|---|---|---|---|
| K | Number of clients | α | Real value of input data |
| $\mathcal{D}_k$ | Private dataset at client | $X_{train}$ | Train dataset |
| $\mathbb{P}$ | Probability distribution | $X_{test}$ | Test dastaset |
| σ | Smoothing parameter | $\mathcal{N}(m, n)$ | Gaussian distribution with mean m and variance n |
| N | Number of samples | η | Learning rate |
| $Dthr$ | Threshold cluster radius | $p_k$ | Representing probability distribution of client 'k' |
| $f_A(X)$ | Activation function | $d(i,j)$ | Normalized euclidean distance between point i and j |
| PDF | Probability density function | $s(i,j)$ | Updated normalized euclidean distance |
| $X_{A_i}$ | Cluster center in the neuron of pattern layer of PNN | $Ru_j$ | $j^{th}$ cluster radius |
| d | Number of independent features | CC | Set of Cluster centers |
| $ρ_e$ | Probability densities of continuous column | $μ_e, Φ_e$ | Weight and Standard deviation of mode 'e' |
| $c_{i,j}$ | Continuous column value | $r_j$ | The row of a dataset |
| b | Server side | | |

All the notation used in the current study is presented in Table 1. Traditionally, FL settings are maintained as follows: local models usually train on the local data respective to each client, {$\mathcal{D}_1,… \mathcal{D}_k$} (refer to Table 1), generated by *k* distinct nodes. Further, data at each node t ∈ *k* is non-i.i.d across the network and is being generated by a distinct distribution $X_t \sim \mathbb{P}_t$ (statistical heterogeneity). Additionally, the number of sample points at each node may significantly vary. These kind of FL settings requires a huge number of communication rounds between clients and the server that suffers the bottleneck in the form of communication overhead. Further, Zhao et al., [34] demonstrated that the invocation of the FedAvg algorithm decreases the quality of FL models. This can result in diverging the global model weights, which leads to poor generalization. There are some additional challenges in the form of network heterogeneity, which implies that some clients in the network are loosely connected to the server. Hence, there is a chance that the updates received from this set of clients would not reach the server and thereby hindering the performance. It resulted in the advent of a new type of learning i.e., one-shot federated learning [11], where there is only one single round of communication between clients and the server. However, as a result, it should not affect the accuracy of the model built. Additionally, in one-shot FL, it is recommended that the number of training samples at each client should be greater than the number of clients to ensure the performance of the FL model comparable to that of the model trained on the central level.

To alleviate privacy concerns, we need to analyze the synthetic dataset. Recently, Generative Adversarial Networks (GANs) [32] occupied a prominent role in generating synthetic datasets by preserving the correlation structure among the features or variables and maintaining the quality. However, there are various challenges, involved in handling tabular datasets viz., (i) handling mixed data types comprising both numerical and categorical features, (ii) dealing with multimodal distribution datasets, and (iii) handling imbalance in categorical features. To overcome these challenges, Xu et al. [9] proposed conditional tabular GAN (CTGAN) by incorporating mode-specific normalization and a conditional vector. However, CTGAN uses single distribution based noise vector which limits the ability to capture the complex dependencies between features. Moreover, it could not capture the true distribution of features. This motivated us to incorporate multiple distributions-based noise vectors to capture complex distributions and increase the probability of capturing the true distribution of features.

Probabilistic neural network (PNN) [33] is a one-pass learning algorithm where training and testing occur simultaneously. It is used for classification problems where the probability density function (PDF) for each class is calculated and computed as the class probability for a given sample. However, the downside of PNN is its computational complexity, which is directly proportional to the size of the training data. Hence, in the case of large datasets, researchers proposed clustering the training data while employing PNN. Clustering algorithms such as K-means, convex clustering, centroid-based clustering, etc., can be used for this purpose. These algorithms usually take multiple iterations to find the optimal cluster centers. Hence, Kasabov and Song [5] proposed an online clustering algorithm i.e., evolving clustering machine (ECM) which takes a single iteration to scan all train data and dynamically compute cluster centers. Here, the number of clusters is controlled by a parameter called Dthr (dynamic threshold radius).

In our current research, owing to the disadvantages of traditional FL, we propose a novel two-stage methodology, where ECM-in the first stage, we employed a modified CTGAN to generate non-i.i.d data. In the second stage, we developed and employed ECM-PNN hybrid classifier to handle high dimensional, voluminous datasets, owing to their one-pass learning capability. The effectiveness of the proposed FL algorithm is tested on tabular datasets taken from finance and medical domains. It is important to note that we conducted an analysis in HFL on tabular datasets and hence the feature space is common among the clients but differs in the sample space.

Major contributions of the current study are as follows:

- Proposed a one-shot FedPNN based classifier by invoking ECM as the pre-processor to PNN.
- Proposed a two-phase methodology for handling tabular datasets from finance and medical domains under one-shot federated settings.
- Proposed a modified CTGAN to generate non-i.i.d synthetic data having multiple distributions with noise.
- Proposed meta-clustering aggregation algorithm for the generalization of global model overcoming the limitations of FedAvg.

The rest of the paper is organized as follows: In Section 2, we presented the literature survey. Section 3 describes the relevant background and Section 4 discusses the proposed methodology. Section 5 describes datasets and Section 6 gives an overview of the experimental setup. Section 7 presents experimental results then Section 8 concludes the paper.

## 2. Literature Survey

In this section, we will discuss the literature review on One-shot algorithms in the context of FL. Zhang et al., [10] proposed three one-shot averaging methods BAVGM (Bootstrap average mixture), achieved comparable performance relative to centralized methods where the data is available at the server. Further, BAVGM turned out to be better than average mixture (AVGM) methods under certain assumptions that the number of samples at the clients is greater than a certain threshold.

Guha et al., [11] proposed a one-shot FL method which is an ensemble of $c \leq d \leq \sqrt{N}$, ( $\because N$ is the number of samples) local models to train the global model. This could be applied to both supervised or semi-supervised learning algorithms. Kasturi et al., [12] proposed a one-shot FL algorithm, named fusion learning where each client sends local dataset distribution parameters and local model parameters to the server. Later, the server then generates the dataset by using all client distributions, labels them using client models then the server trains the global model on a generated dataset.

Shin et al., [13] proposed a privacy-preserving data augmentation technique by employing a mix-up method under one-shot FL settings and named XorMixFL to handle non-i.i.d data. The server then decodes the encoded seed samples by employing the XOR operator. Here, these seed samples are uploaded by the clients and adds them to the base samples at the server to iteratively construct a balanced dataset until it converges. It is reported that the global model trained on reconstructed data performed better than the traditional FL settings.

Zhou et al., [14] proposed a distilled one-shot federated learning, named DOSFL, where the server trains on distilled training data sent by available clients. Qinbin et al., (2021) proposed a practical one-shot federated learning algorithm for cross-silo settings named Federated knowledge transfer (FedKT), which uses a knowledge transfer technique for the clients to learn from the local client model and the server uses these models as an ensemble to make the predictions on the public data at the server.

Salehkaleybar et al., [15] analyzed problems of statistical optimizations of convex loss landscapes in one-shot methods, and under communication constraints. They proposed a method called multi-resolution estimator clustering with communication cost of log (mn) bits per transmission (MRE-C-log), under constrained estimation error, which is able to reach order optimal up to poly-logarithmic factor.

Dennis et al., [16] proposed a one-shot Federated clustering technique (K-FED), a variant of distributed K-means clustering algorithm [27], which leverages data heterogeneity and mitigates data distribution for clustering. Eren et al., [17] proposed a first-of-its-kind recommendation system using collaborative filtering in one-shot FL (FedSPLIT) where each client locally trains a recommendation model using collaborative filtering and global bias which is received from the server. Later, it sends the local pattern and bias of these models back to the server. The server uses joint non-negative matrix factorization [4] to aggregate the client updates.

Zhang et al., [18] proposed a data-fee one-shot FL (DENSE) where the server trains an auxiliary generator by using an ensemble of local model updates from the clients after that the final global model is trained. This is followed by the generation of synthetic data by the trained generator and an ensemble of client models using knowledge distillation. Lee et al., [19] proposed a one-shot FL where linear unbiased estimates of multi-task linear (MTL) regression at the clients are uploaded to the server. Then, it linearly combines these estimates and produces improved MTL estimates. Fusion learning at the server is performed by collecting the estimates in each iteration and aggregating the updated estimate to find the final MTL.

Hoech et al., [20] proposed a one-shot FL, Federated auxiliary differential private (FedAUXdf) where each client trains two models i.e., graph regularized logistic classification and a proposed differentially private model. The server trains the full global model using aggregated weight updates of FedAUXdf model, from the clients and public distilled data, labeled using certainty scores of local logistic classification trained on clients. Rjoub et al., [21] integrated a model-free reinforcement learning component into a one-shot FL (OSFL) algorithm where local models are aggregated at the server and clients can choose to label their samples either automatically or request the true label for a one-shot learning component.

Garin et al., [22] provided upper bounds on the local MSE and biases in a very general setting of empirical risk minimization (ERM). They proposed Federated estimation with statistical correction (FESC) which aggregates the local estimations based on a minimization of an upper bound of the MSE. They also addressed the minimum sample size constraint of the clients in the one-shot FL scenario. Garin and Quintana [23] proposed FESC method for the effective sample size distribution of the clients. They showed that the mean number of samples per node must be at least equal to the number of nodes in the FL training for its performance comparable to that of centralized training.

Wang et al., [24] proposed a novel, first-of-its-kind Data-free diversity-based ensemble selection framework (DeDES) for selecting strong ensemble teams from the machine learning model market, these models are trained by one-shot FL. They proposed a technique for selecting a better representative model inside a cluster to improve the performance of the final model.

Humbert et al., [25] presented an institute one-shot FL method named Federated conformal prediction based on quantiles of quantiles (FedCP-QQ). The prediction scores include differentially private data points and their corresponding prediction scores for each client. These scores are arranged in the quantile range, and the model selects local empirical quantiles in a privacy-preserving manner. Then, they are sent to the server where the aggregation of the quantile of these quantiles is executed.

Armacki et al., [26] proposed a family of one-shot federated clustering algorithms based on simple inference and averaging schemes with K-means and convex clustering methods. The theoretical results showed that the order-optimal mean squared error (MSE) guarantees results compared to central learning in terms of space complexity.

Our current study is different from the literature in the following way:

- Proposed One-shot FedPNN based classifier.
- To the best of our knowledge, our methodology is the first-of-its-kind, employing ECM invoked with PNN in FL scenario to further speed up the process.
- We incorporated two distributions in CTGAN and proposed a modified CTGAN.

## 3. Preliminaries

In this section, we discuss the employed architectures for classification and synthetic data generation.

### 3.1 Evolving Clustering Method

Evolving clustering method (ECM) [5] is an online dynamic clustering method, a one-pass learning algorithm that uses normalized euclidean distance to dynamically find the cluster centers. It was applied to anomaly detection [4], two spiral problems [5], etc.

ECM starts by considering the first sample as the cluster center, with a radius initialized to zero. Whenever a new sample comes, then the euclidean distance between the point and all of the existed centers is calculated individually and the minimum is considered for further evaluation. The current sample either falls into one of the existing cluster centers or forms a new cluster which is decided by using dynamic threshold radius (Dthr) value. If the minimum distance is less than the two times of Dthr value then the new sample is added to that particular cluster, and then both the cluster center and radius are updated accordingly. Otherwise, a new cluster is formed by considering the current sample as the new cluster center. This process is repeated until all samples are completely scanned.

## 3.2 Probabilistic Neural Network

PNN [33] is a single pass learning based feed forward neural network used for classification and pattern recognition problems such as credit card fraud detection, character recognizing, prediction of leukemia, and embryonal tumour of the central nervous system [7], etc. The architecture of PNN is depicted in Fig. 1.

PNN majorly comprises four layers (i) input layer, (ii) pattern layer, (iii) summation layer, and (iv) output layer. The input layer transfers the input test data to the pattern layer. Each neuron in the pattern layer contains a labeled training sample. The hidden neuron computes the euclidean distance between the test sample and the center of the neuron and applies a Gaussian, Logistic, or Cauchy activation function. The summation layer then calculates the weighted sum of probability estimates corresponding to each class, and the output layer compares the probability estimates of each category and outputs the most probable class. It uses Parzen's approach to estimate the parent probability density function (PDF) of a given set of training data. Thus, the estimated PDF would asymptotically approach the Bayesian optimal by minimizing the "*expected risk*" which is regarded as the "*Bayes strategies*".

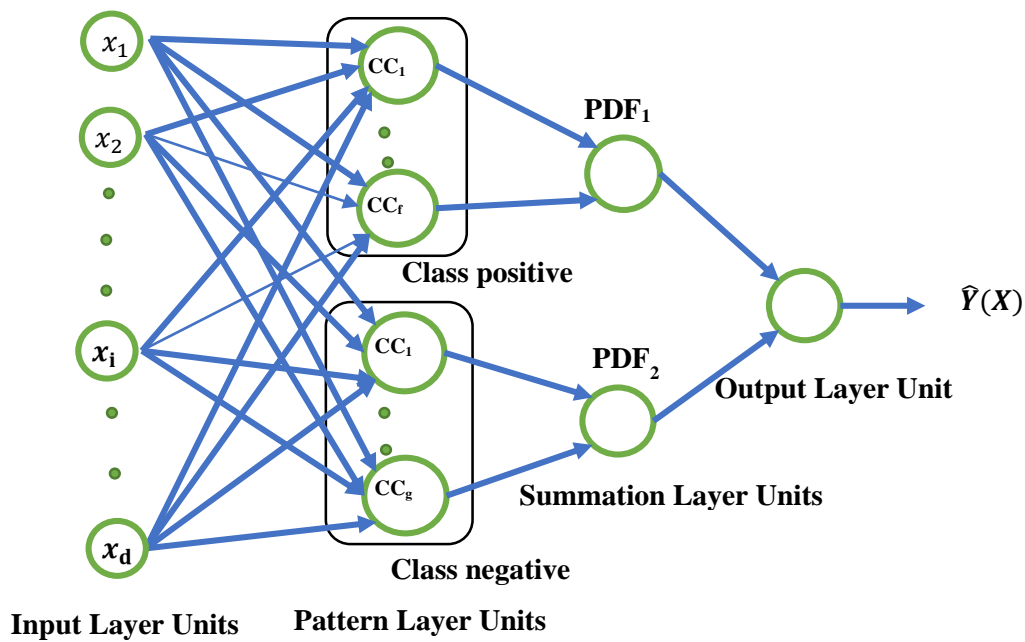

**Fig. 1:** Architecture of PNN

In our current study, we adopted the Gaussian activation function is given in Eq. 1.

$$f_A(X) = \frac{1}{n(2\pi)^{d/2}\sigma^d} \sum_{i=1}^{n} exp\left[-\sum_{j=1}^{d}\left(X_j - X_{A_{i,j}}\right)^2 \Big/ 2\sigma^2\right] \quad (1)$$

where, n is the number of train samples of a class, σ = smoothing parameter, and d is dimensionality of measurement space or number of features in the dataset.

The computational complexity of PNN is relative to the size of the training data. Hence, researchers paved us to utilize the clustering algorithms such as K-Means, Convex clustering, ECM etc., to decrease the complexity. It is important to note that after performing the clustering, the cluster centers will reside in neuron of the pattern layer.

### 3.3 Overview of Conditional Tabular GAN

Conditional Tabular GAN (CTGAN) is a GAN-based method that can synthesize real tabular data distribution. The architecture is depicted in Fig. 2. The following are a few important advantages of CTGAN : (i) can handle both continuous and discrete features, (ii) has the ability to model non-gaussian and multimodal distribution, and (iii) can handle a severe imbalance in continuous features.

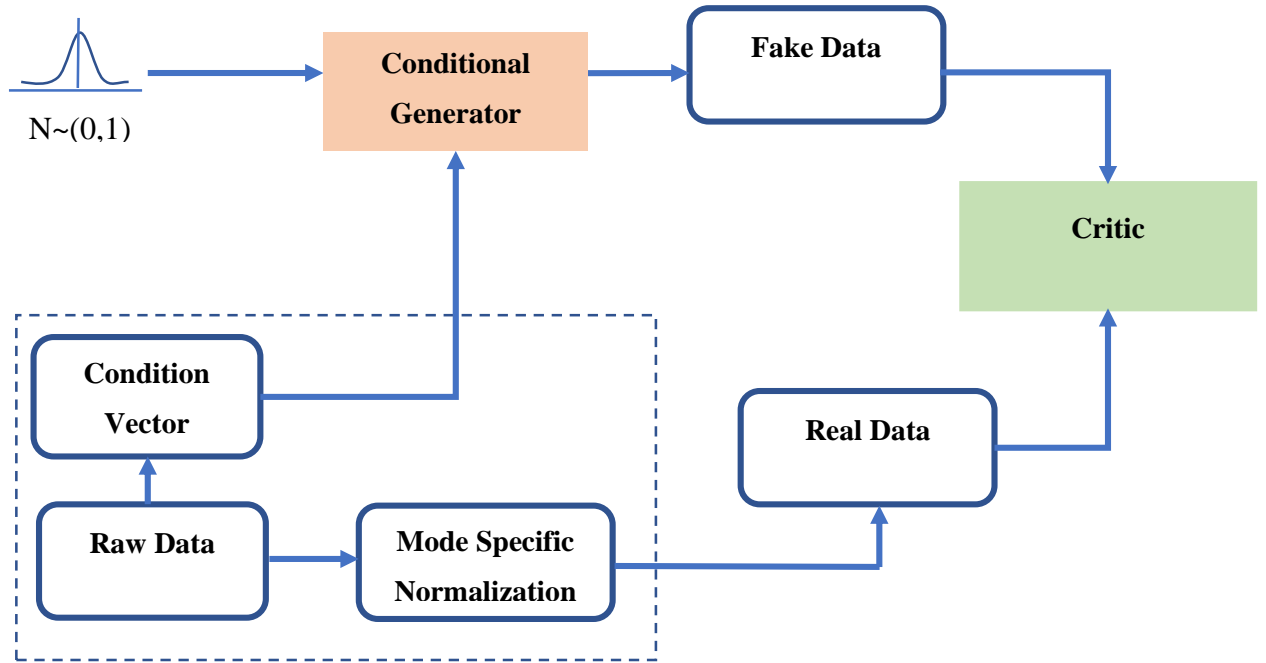

**Fig. 2:** CTGAN architecture (adapted from Moon et al. [30])

CTGAN mainly consists of three key elements namely, (i) Conditional Vector, (ii) Generator loss, and (iii) Training by sampling. It consists of two architectures viz., generator and critic. During training, the generator can produce the output of any vector. CTGAN handles both the numerical and categorical features as follows: (i) Incorporate the mode-specific normalization where each column is trained independently. It estimates the number of modes ($m_i$) by using the Variational gaussian mixture model (VGM) for each $F_i$ and generates an encoded vector that can be used in place of the original values. (ii) Owing to the behaviour of GAN

Page | 8

which gets biased towards generating only the most frequent categories and thereby ignoring the less frequent classes. Therefore, while handling the categorical features, it adapts a conditional vector where multiple conditions can be specified to handle specific categorical columns. Hence, this is regarded as the conditional generator. Here, these conditions can relate to multiple categorical features thereby making CTGAN to handle single or multiple categorical features simultaneously.

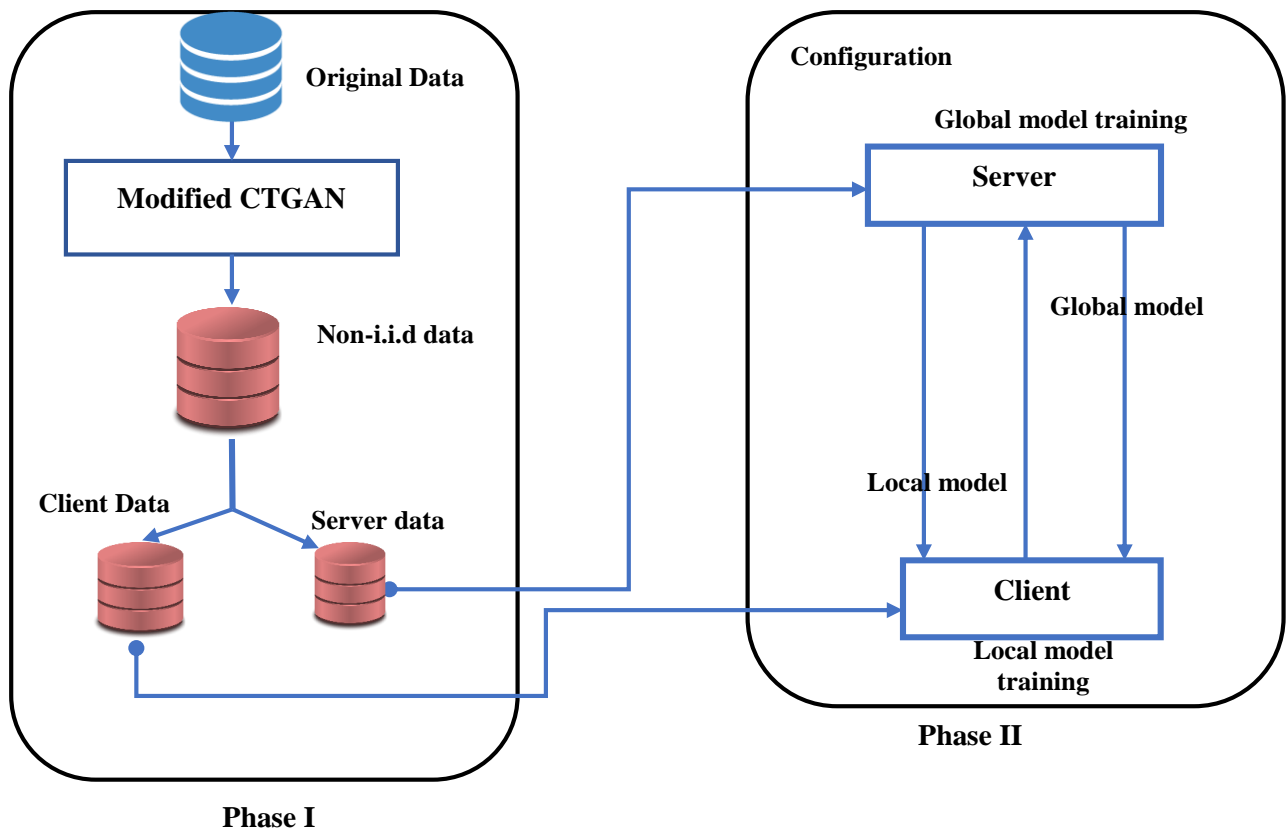

**Fig. 3:** Flow diagram of proposed methodology

## 4. Proposed Methodology

In this section, we will discuss the proposed framework which is depicted in Fig. 3, and the algorithm is given Algorithm 1. It comprises two phases, namely, Phase-I where the non-i.i.d synthetic data is generated, which ensures privacy and removes uniquely identifiable information from the real dataset. Thereafter, Phase-II begins where FedPNN is employed with the meta-clustering aggregation algorithm. The above two phases are discussed in detail below:

## PHASE-I : Non-i.i.d Synthetic data generation

The following is performed in the non-federated settings as follows:

- A synthetic non-i.i.d dataset is generated by the modified CTGAN.
- The thus generated synthetic non-i.i.d dataset is shared among *k* clients by employing a random sampling technique and then we proceed to Phase-II.

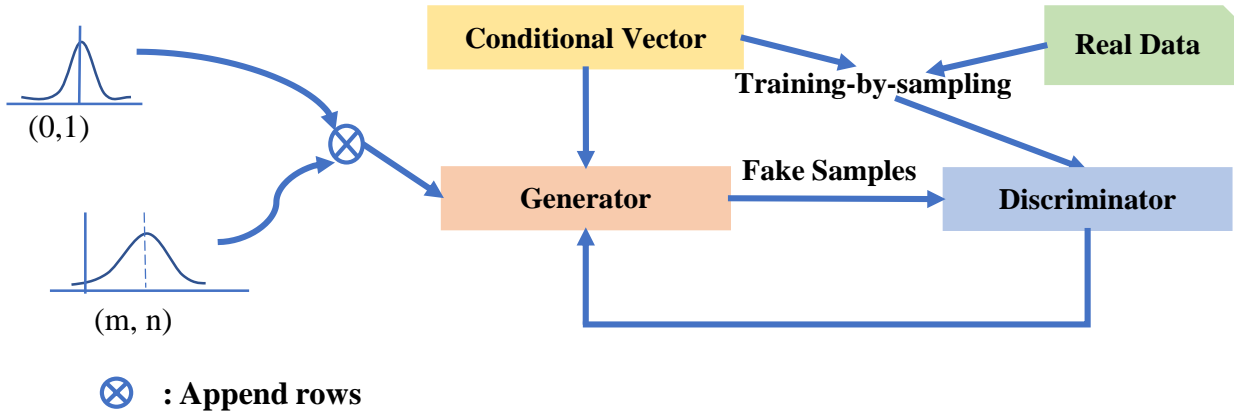

**Fig. 4:** Modified CTGAN

**Step a: Modified CTGAN**

In the proposed architecture, we proposed and employed a modified CTGAN (refer to Fig. 4) which is different from Vanilla CTGAN in the following way:

- Vanilla CTGAN incorporates a single distribution to generate the input noise from randomly sampled data. However, in the modified CTGAN, we employed noise generation from two distributions (refer to Fig. 3) to overcome the challenges such as capturing complex dependencies between features that are faced by single distribution based noise. The employed distributions are: (i) standard normal $\sim(0,1)$, and (ii) normal distribution $N\sim(0.2,2)$.

Once, the random noise is generated from multiple distributions then it is fed as an input to the Generator. The conditions corresponding to categorical columns should be preserved in a conditional vector to generate specific target samples. The generator network in GAN considers both the random noise and conditional vector information to produce the output. It is to be noted that, we employed two different CTGANs (refer to Fig. 4) one for positive sample generation (referred to as $CTGAN_{pos}$) and the other is $CTGAN_{neg}$ for negative sample generation. Doing so leverages us to generate as many as samples specific to a single class. As we know that, handling imbalance is very critical in the One-shot FL settings because the underlying model may get biased towards one class. To circumvent these challenges we employed two CTGANs each for a class.

Once the dataset is generated, its quality is assessed by using the two different synthetic dataset metrics namely, (i) Koglomorinov Smirnov (KS) complement test score to assess the similarity between real samples and synthetic samples, and (ii) mean correlation symmetry test score (CStest score) to assess how much correlation structure is preserved between real and synthetic datasets. Since, we employed two CTGANs, the generated two datasets are combined and the quality of the synthetic data is assessed. Once, the training is completed and the quality metrics are satisfied then the next step begins.

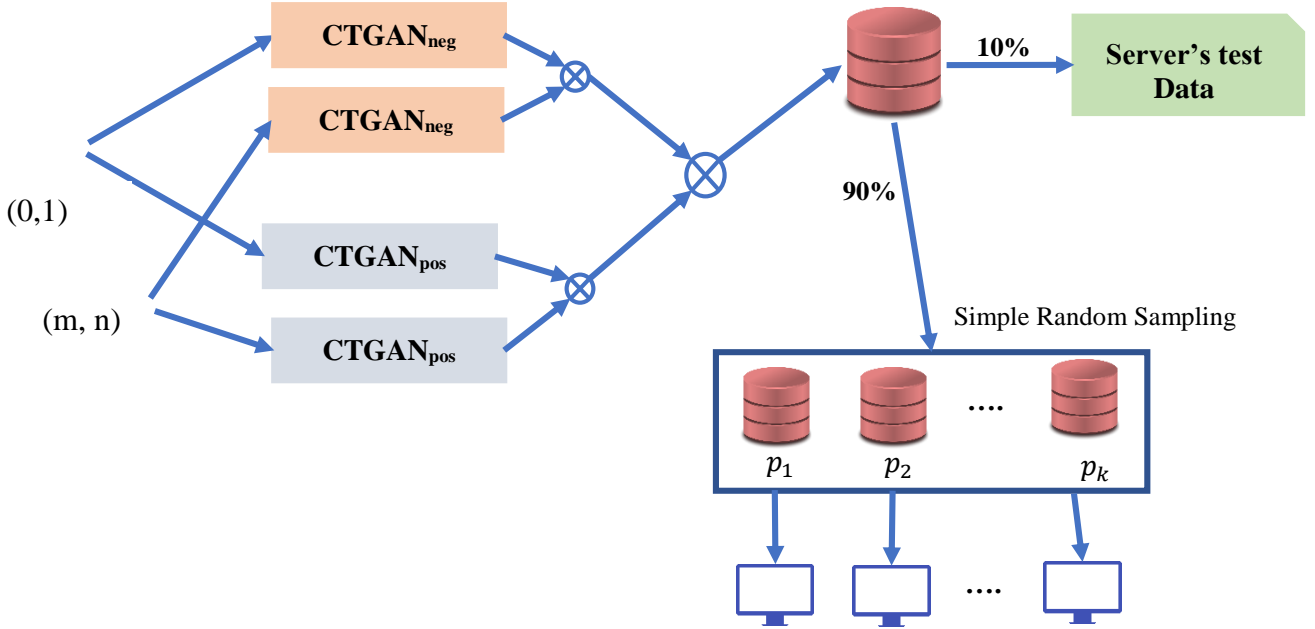

**Fig. 5:** Non-i.i.d synthetic data generation and data sharing among the clients

**Step b: Data Sharing**

Once, the synthetic dataset is generated, it is shared among clients and the server as follows:

- We adopted *b% :(100-b)%* proportion between the server and client nodes by employing the stratified random sampling method. In our one-shot FL settings, we reserved *b* % of the synthetic dataset at the server node for the server side evaluation. This helps in evaluating the performance of the meta-clustering aggregation principle, discussed later on.
- The rest of the (*100-b*)% of the dataset is shared among clients via simple random sampling which ensures the non-i.i.d nature of the dataset among clients. Further, each client receives the same proportion of samples of the classes as present in the original data and thus meets the real-world requirements.

In our approach, we fixed *b* to be 10 and hence the rest 90% of the dataset is shared among clients.

## PHASE II: FedPNN

In this phase (refer to Fig. 6), we employed PNN, a one pass learning algorithm that is preceded by a modified ECM algorithm. The performance of the modified ECM is also susceptible to Dthr, which denotes the cluster radius, and is a hyperparameter that determine the number of cluster centers.

The following are the critical steps involved:

- All the required parameters are broadcasted across the cluster.
- PNN models are locally evaluated at each client and then the centers are collected back at the server.
- Then, the meta-clustering algorithm is employed that involves clustering of the centers obtained by various clients by invoking ECM again and thus serves as an aggregation algorithm.
- Meta centers are passed back to the client and then evaluated on the updated local models at each client.

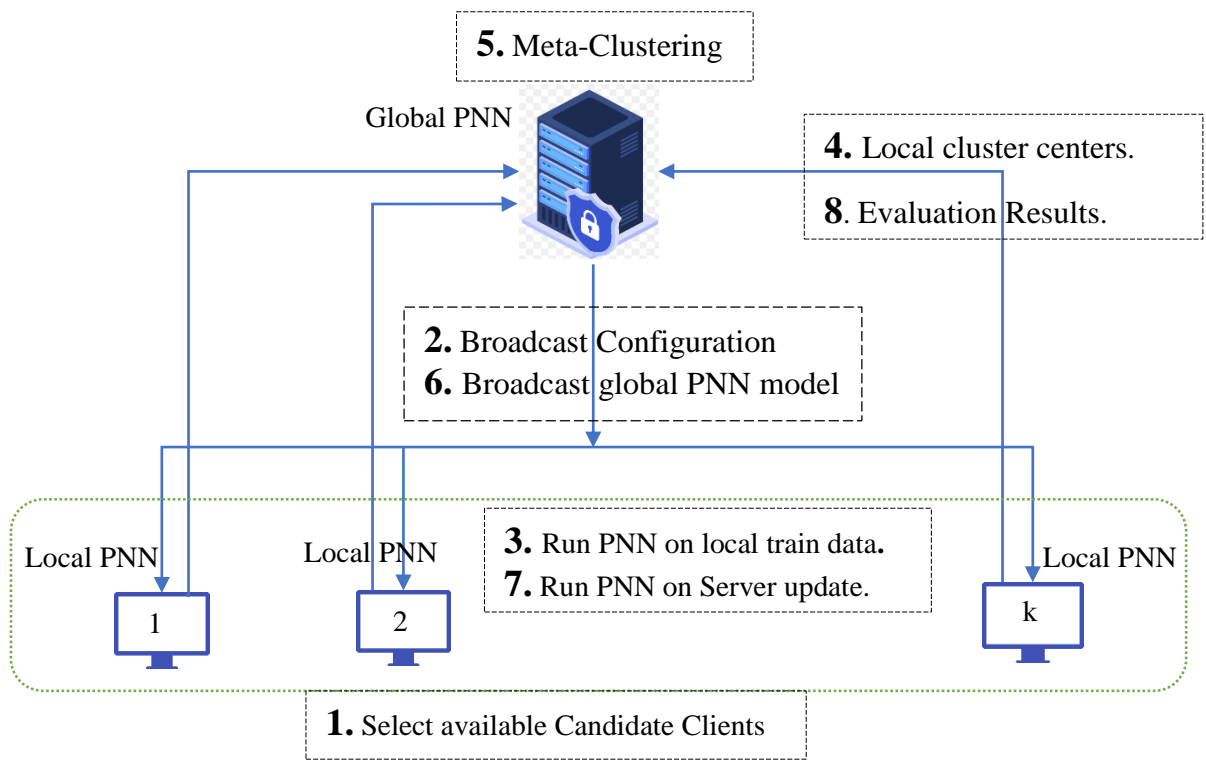

**Fig. 6:** Proposed one-shot FL Architecture

**Step a: Broadcasting information**

In the first step, as mentioned in Phase-I, we reserved *b* % of the generated synthetic dataset on the server side. Hence, we need to maintain Dthr for both the clients and server individually. All these parameters are broadcasted to all the nodes in the federated environment.

**Step b: Local PNN training**

We know that the (*100-b*) % dataset is reserved for clients and among clients simple random sampling dataset is employed thereby each client receives {$p_1, p_2, ..., p_k$}. Now, within each client, we maintained the train-test ratio of 80%:20%, and adopted stratified random sampling. By using client Dthr, we employed modified ECM (refer to Algorithm 2) to compute the cluster centers.

Modified ECM is different from ECM in the following way: It tracks the frequency of the points belonging to each of the clusters which are later used to label the cluster either as positive or negative class.

$$d(i,j) = \left\| x_i - C_{c_j} \right\|, j = 1, 2, ... n \quad (2)$$

$$\|x - y\| = \left( \sum_{i=1}^{d} |x_i - y_i|^2 \right)^{1/2} \bigg/ d^{1/2}, \quad x, y \in R^d \quad (3)$$

**Algorithm 1: FedPNN**

**Server executes:**

**Input**: K, $X_{train}$, $X_{test}$

**Output:** AUC at the server and clients

1. **CC** ← [ ]$_{*XK}$
2. freq ← [ ]$_{pXK}$
3. **for** k := 1 to K **do**
    //client update function to receive results from the client k
4.     $CC_k$, $freq_k$ ← clientUpdate$_k$(Dthr, p)
5.     **CC**.append($CC_k$), freq.append($freq_k$)
6. **end** for
7. CC ← Meta_clustring(**CC**, freq)
8. Z ← group(CC) // grouping the $X_{train}$ based on class.
9. score ← **EPNN**(Z, $X_{test}$)
10. **broadcast** CC to K clients

**Client executes:**

**Input:** $D_k$: private dataset at the client

**Output:** AUC

1. $X_{train}$, $X_{test}$ ← $D_k$
2. Normalize ($X_{train}$, $X_{test}$)
    // server invokes this method to receive the result
3. **function** clientUpdate(Dthr, p)
4.     **CC** ← **ECM**($X_{train}$, Dthr, p)
5.     AUC ← **EPNN**(CC, $X_{test}$)
6.     **return CC** // returns the update to the server.
7. **CC** ← Server broadcast
8. AUC←**EPNN**(CC, $X_{test}$)
9. **return** CC, AUC

## Algorithm 2: Modified ECM

**Input:** $X_{train}$, n, Dthr, p

**Output:** **CC**, freq

1. $X_{train}, y_{train} \leftarrow X_{train}$
2. $X_{norm} \leftarrow$ Normalize ($X_{train}$), **CC** $\leftarrow$ [ ]$_{*xd}$, freq $\leftarrow [0,0,...,0]_{xp}$ // number of classes equals 2
3. **CC**.append($X_{norm}$[1]), $Ru_1 \leftarrow 0$
4. freq[1, $y_{test}$[1]] $\leftarrow$ freq[1, $y_{test}$[1]] + 1 // $y_{test}$[*] can take value 1 or 2 for binary classification.
5. **for** j := 2 to n **do**
6.     $x \leftarrow X_{norm}[j], l \leftarrow$ len(**CC**)
7.     **compute** $d(x,h)$, $h = 1,2, ..., t$     (refer to Eq. (4))
8.     $d(x,m) \leftarrow \min_k d(x,h)$, $h = 1,2, ...,t$
9.     **if** $d(x,m) \leq Ru_m$ **then**
10.         freq[m, $y_{test}$[j]] $\leftarrow$ freq[m, $y_{test}$[j]] + 1
11.         **return to step 4**
12.     **else**
13.         $s(x,h) <- d(x,h) + Ru_h$, $h <- 1,2,..t$ (refer to Eq. (4))
14.         $s(x,a) <- \min_h s(x,h)$, $h <- 1,2,..t$ (refer to Eq. (5))
15.     **end if**
16.     **if** $s(x,a) > Dthr$ **then**
17.         **CC**.append($X_{norm}$[j]), $Ru_{l+1} \leftarrow 0$, freq.append ([[0], [0]])
18.         freq[m, $y_{test}$[j]] $\leftarrow$ freq[m, $y_{test}$[j]] + 1
19.         **return to step 4**
20.     **else**
21.         $Ru_a \leftarrow s(x,a)/2$, $e \leftarrow$ **CC**[a]
22.         temp $\leftarrow d(x,e)$
23.         ratio $\leftarrow abs(temp - Ru_a)/temp$
24.         **CC**[a] $\leftarrow$ **CC**[a] + $(x - e) *$ ratio
25.         freq[a, $y_{test}$[j]] $\leftarrow$ freq[a, $y_{test}$[j]] + 1
26.     **end if**
27. **end for**
28. **return CC**, freq

The training of the modified ECM is described below:

- It starts by creating the first cluster center Cc by simply considering the first data point. For each cluster, there is a cluster center $Cc_k$ and the corresponding radius $Ru_k$. It employs normalized euclidean distance (refer to Eq. 2 and Eq. 3) which is computed for a given point and existing cluster centers.
- Now, the cluster center which is in the minimum distance (refer to Eq. 4) is considered for deciding whether the current data point belongs to a cluster or forms a new cluster. This is decided by comparing the minimum distance with Dthr.

- It is important to note that the minimum distance refers to the additive value of both normalized euclidean distance and radius of the corresponding cluster.
- Now, thus obtained minimum distance is compared with the Dthr condition as follows:
    - If $s(i,a) > 2 \times Dthr$, then a new cluster center with the current data point is formed,
    - If $s(i,a) \leq 2 \times Dthr$, then the data point is to be allotted to that particular cluster $a$ and then update the cluster center, the corresponding radius, and class frequency vector.

$$s(i,a) = \{ \forall j = 1,2,\dots,n\, , \min(d(i,j) + Ru_j)\} \quad (4)$$

**Algorithm 3: PNN**

**Input:** CC, $X_{test}$, $y_{test}$

**Output:** score

1. u ← len($X_{test}$)
2. $y_{pred}$ ← [ ]$_{1 \times u}$
3. **for** i := 1 to u **do**
4.     **for** j := 1 to len(**CC**) **do**
5.         $A_j$ ← 0,
6.         m ← len(**CC**$_j$)
7.         **for** h := 1 to m **do**
8.             temp ← d($X_{test}$[i], **CC**$_j$[h])    ( refer to Eq. 2)
9.             $A_j$ ← $A_j$ + Gaussian(temp)    ( refer to Eq. 1)
10.         **end for**
11.         $A_j$ ← $A_j$ / m
12.     **end for**
13.     $y_{pred}$[i] ← indexOfMax([$A_1, A_2, \dots, A_l$])//this function returns the index of the max value
14. **end for**
15. score ← computeAUC ($y_{test}$, $y_{pred}$)
16. **return** score

Finally, by using class frequency vector, the dominating class is assigned to the cluster. In the case of having the same frequencies, arbitrarily one of the classes is assigned to that cluster. Once, the ECM is parsed on the entire training dataset results in the generation of cluster centers. All these cluster centers resided at the pattern layer of PNN (refer to Algorithm 3), where each test sample from local data is passed on to the neurons present in the pattern layer and undergoes the respective activation function (i.e., Gaussian, Logistic, or Cauchy). All of these activation functions results in a higher value if a neuron has a higher similarity to the test sample. In our current study, we employed the Gaussian activation function. Then, all the outputs from the pattern layer are gathered into the summation layer and PDFs for each class involved in the classification process is computed. Thereafter, the output is predicted at the output layer of PNN.

Upon the completion of local PNN training at each client, the cluster centers obtained from ECM are sent to the server, where the meta-clustering aggregation is executed.

## Algorithm 4: Meta-clustering using modified ECM at the Server

**Input:** X, Y, n, Dthr, p

**Output:** **CC**, freq

**CC** ← [ ], freq ←[[[0], [0]]]$_{*xd}$ // number of classes equals 2

1. **CC**.append(X[1]), $Ru_1$ ← 0
2. freq[1,1] ←freq[1,1]+**Y**[1,1]  // **Y**[1,*] can take value 1 or 2 for binary classification.
3. freq[1,2] ←freq[1,2]+**Y**[1,2]
4. **for** j := 2 to n **do**
5.     x ← X$_{norn}$[j], l ← len(CC)
6.     compute $d(x, h)$, $h = 1,2,..l$ using Eq (2)
7.     $d(x, m) \leftarrow \min_{k} d(x, h)$, $h = 1,2,..l$
8.     **if** $d(x, m) \leq Ru_m$ **then**
9.       freq[m,1] ←freq[m,1] + **Y**[j,1]
10.       freq[m,2] ←freq[m,2] + **Y**[j,2]
11.       **return to step 4**
12.     **else**
13.       $s(x, h) = d(x, h) + Ru_h$, $h = 1,2,..l$ using Eq 4
14.       $s(x, a) = \min_{h} s(x, h)$, $h = 1,2,...l$ using using Eq 5
15.     **end if**
16.     **if** $s(x, a) > Dthr$ **then**
17.       **CC**.append(X$_{norn}$[j]), $Ru_{l+1}$ ← 0, freq.append ([[0], [0]])
18.       freq[l,1] ←freq[l,1] + **Y**[j,1]
19.       freq[l,2] ←freq[l,2] + **Y**[j,2]
20.       **return to step 4**
21.     **else**
22.       $Ru_a \leftarrow s(x,a)/2$, $e \leftarrow$ **CC**[a]
23.       $temp \leftarrow d(x, e)$
24.       ratio ← $abs(temp - Ru_a)/temp$
25.       **CC**[a] ← **CC**[a] + $(x - e) *$ ratio
26.       freq[a,1] ←freq[a,1] + **Y**[j,1]
27.       freq[a,2] ←freq[a,2] + **Y**[j,2]
28.     **end if**
29. **end for**
30. **return CC,** freq

**Step c: Meta-clustering aggregation algorithm**

In the first step, as mentioned in Phase-I, we reserved $b$ % of the generated synthetic dataset on the server side. After training the local PNN model at each client, cluster centers are obtained to the server. As we know that, due to the heterogeneous number of clusters, one cannot employ the Federated average (FedAvg) or its successor algorithms. Hence, we proposed meta-clustering, where we employed ECM to find meta-cluster centers is given in Algorithm 4. It is to be noted that, the meta-cluster centers are computed on the cluster centers obtained from all the clients. Using these meta-cluster centers, the PNN model is evaluated at the server side.

**Step d: Global training**

The meta cluster centers which are obtained from step c of Phase-II, are broadcasted to all the clients and are used to update the local models at each client and then these meta clusters are fed into the pattern layer of the local PNN model. The same test process which is discussed in local PNN training (Step b of Phase II) is performed by using meta clusters. Later, AUC is evaluated for each client.

## 5. Dataset Description

In the current study, we considered four benchmark datasets related to banking and medical domains that are binary classification datasets (refer to Table 2). It should be noted that 3 out of 4 datasets except Wilcoxon Breast Cancer are highly imbalanced datasets and all of them are having only numerical features. Originally, OVA Omentum dataset features consisted of 10,936, but we employed t-statistic based feature selection and selected the top 500 features for our analysis.

For each benchmark dataset, synthetic datasets were generated using modified CTGAN and the minority class is oversampled. The description of the synthetic data is presented in Table 3.

**Table 2:** Description of Benchmark datasets.

| Dataset | Data points | Features | Class Distribution ratio | |
|---|---|---|---|---|
| | | | Negative Class (%) | Positive Class |
| Breast Cancer | 699 | 11 | 65.50 | 34.50 |
| OVA Omentum | 1545 | 10936 | 95.00 | 5.00 |
| Credit Card Fraud | 2,84,807 | 32 | 99.83 | 0.17 |
| Polish Bankruptcy | 10,172 | 65 | 96.10 | 3.90 |

**Table 3**: Description of generated synthetic data of Benchmark dataset

| Dataset | Data points | Features | Class Distribution ratio | |
|---|---|---|---|---|
| | | | Negative Class (%) | Positive Class |
| Breast Cancer | 916 | 11 | 50.00 | 50.00 |
| OVA Omentum | 2936 | 10936 | 50.00 | 50.00 |
| Credit Card Fraud | 3,98,237 | 31 | 71.00 | 29.00 |
| Polish Bankruptcy | 19,544 | 65 | 50.00 | 50.00 |

## 6. Experimental Setup

The following is the federated experimental setup utilized for conducting the experiments: one server and two clients. All the nodes are having Intel i7 6[th] generation processors with 32 GB RAM. For FL implementation, we utilized the Flower framework [29] to conduct the experiment which is developed in Python language and supports scaling to many distributed clients. It is also flexible for adopting a new proposed approach with low engineering overhead.

**6.1 Evaluation Metrics**

In the current study, we employed two metrics for evaluating the quality of the synthetic dataset (i) Complement of the Kolmogorov-Smirnov test (KSComplement), (ii) Correlation similarity test score (CStest). Additionally, the performance of FedPNN is evaluated by AUC. The metrics are explained below:

**6.1.1 KSComplement**

It measures the similarity of a given column in the real and synthetic datasets. It uses the Koglomorinov-Smironov statistic (KS statistic), which is calculated by converting the numerical column into its cumulative distribution frequency (CDF). The maximum difference beteen the two CDFs is known as the KS statistic, which lies between [0,1]. The KSComplement metric, mathematically defined in Eq. 6, also lies within the range of [0,1]. The higher the KSComplement the better the similarity is between real and synthetic columns.

$$\text{KSComplement test Score} = 1 - \sup_x |F_i(x) - F_j(x)| \qquad (6)$$

where, $\sup_x$ is the supremum of set of distances. $F_i$ is the Cumulative Distribution Function (CDF) of real data continuous columns likewise $F_j$ is the CDF of the same continuous column in synthetic data. Mean KSComplement test score is the mean of test scores of all the independent variables of a dataset.

**6.1.2 CStest**

It measures the correlation similarity between two numerical columns of the real and synthetic datasets. In our current study, the CStest score is presented in Eq. 7.

$$\text{Correlation Similarity test Score (A, B)} = 1 - \frac{|S_{A,B} - R_{A,B}|}{2} \qquad (7)$$

where, $S_{A,B}$ and $R_{A,B}$ are the correlation coefficient [28] of A and B, a pair of continuous columns, on synthetic and real data respectively.

**6.1.3 AUC**

While handling imbalanced datasets, AUC is proven to be a robust measure which is an average of specificity and sensitivity. It is mathematically represented as presented in Eq. (8).

$$AUC = \frac{(Sensitivity + Specificity)}{2} \qquad (8)$$

where,

$$Sensitivity = \frac{TP}{TP + FN} \qquad (9)$$

and

$$Specificity = \frac{TN}{TN + FP} \qquad (10)$$

where TP is a true positive, FN is a false negative, TN is a true negative, and FP is a false positive.

## 7. Results and Discussion

All the hyperparameters used for the synthetic dataset generation and for the current study are presented in Table 4. In this section, we will discuss the following: (i) synthetic dataset metric analysis, (ii) AUC analysis after local training, and meta-clustering. Further, we conducted a sensitivity analysis to discuss the effect of Dthr over the performance in the one shot FL settings in two different ways viz., (i) by varying client Dthr and keeping server Dthr constant, and (ii) by varying server Dthr and keeping client Dthr constant.

**Table 4**: Hyper parameter table used in the current study

| Dataset | GAN Architecture | | Dthr | |
|---|---|---|---|---|
| | **Negative Class** | **Positive Class** | **Clients$_1$ / Client$_2$** | **Server** |
| **Breast Cancer** | GEN architecture (32,16) <br> DIS architecture (16,32) <br> Epochs = 80 | GEN architecture (16,32) <br> DIS architecture (32,16) <br> Epochs = 50 | 0.19 | 0.17 |
| **OVA Omentum** | GEN architecture (256,128,64) <br> DIS architecture (64,128,256) <br> Epochs = 50 | GEN architecture (256,128,64) <br> DIS architecture (64,128,256) <br> Epochs = 20 | 0.13 | 0.07 |
| **Credit card fraud** | GEN architecture (16,32) <br> DIS architecture (32,16) <br> Epochs = 50 | GEN architecture (16,32) <br> DIS architecture (32,16) <br> Epochs = 150 | 0.17 | 0.13 |
| **Polish Bankruptcy** | GEN architecture (64,32,16) <br> DIS architecture (32,128) <br> Epochs = 50 | GEN architecture (64,32) <br> DIS architecture (256,256) <br> Epochs = 30 | 0.11 | 0.05 |

* GEN : Generator ; DIS : Discriminator

## 7.1 Comparative Analysis of Synthetic Data metrics

In this section, we compared the performance of the proposed modified CTGAN with that of vanilla CTGAN based on KSComplement and CStest scores. In our settings, we employed two CTGANs each for the positive and negative classes, respectively. We presented the mean KSComplement and mean CStest scores in Table 5 and Table 6, respectively. Importantly, it is desired to have a higher KS complement score and CStest score for a better synthetic dataset.

Table 5 shows that in 3 out of 4 datasets, Vanilla CTGAN obtained better KSComplement scores than modified CTGAN. This demonstrates that the synthetic dataset generated by Vanilla CTGAN is more similar to the original dataset when compared to that of modified CTGAN. Interestingly, in the Breast cancer dataset, modified CTGAN obtained a mean KSComplement score of 0.882 which is 1.9% greater than that of Vanilla CTGAN. However, in the Polish Bankruptcy, modified CTGAN obtained just 29.3% which is too less than the KSComplement score of Vanilla CTGAN. Overall, in terms of the KSComplement score, Vanilla CTGAN turned out to be better.

**Table 5**: Mean KSComplement scores of the generated synthetic data

| Dataset | Vanilla CTGAN | | | Modified CTGAN | | |
|---|---|---|---|---|---|---|
| | Negative Class | Positive Class | Mean | Negative Class | Positive Class | Mean |
| **Breast Cancer** | 0.868 | 0.859 | 0.8635 | 0.877 | 0.886 | **0.882** |
| **OVA Omentum** | 0.728 | 0.704 | **0.716** | 0.691 | 0.661 | 0.676 |
| **Credit card fraud** | 0.914 | 0.793 | **0.8535** | 0.920 | 0.688 | 0.804 |
| **Polish Bankruptcy** | 0.6122 | 0.929 | **0.7706** | 0.262 | 0.325 | 0.293 |

From Table 6, it is demonstrated that modified CTGAN obtained a better CStest score ensuring that it captured the correlation structure of the original dataset. Interestingly, Vanilla CTGAN obtained almost similar CStest scores in both the Breast cancer and OVA Omentum datasets respectively. It is important to note that even though modified CTGAN failed miserably to replicate the similarity of the original dataset (see Table 5), yet it captured correlation structure by 13% better than Vanilla CTGAN. Overall, in terms of the CStest score, modified CTGAN turned out to be better than Vanilla CTGAN.

**Table 6**: Mean CStest scores of the generated synthetic data

| Dataset | Vanilla CTGAN | | | Modified CTGAN | | |
|---|---|---|---|---|---|---|
| | Negative Class | Positive Class | Mean | Negative Class | Positive Class | Mean |
| **Breast Cancer** | 0.860 | 0.896 | **0.878** | 0.868 | 0.888 | **0.878** |
| **OVA Omentum** | 0.912 | 0.920 | **0.916** | 0.911 | 0.920 | **0.916** |
| **Credit card fraud** | 0.981 | 0.856 | 0.918 | 0.980 | 0.860 | **0.920** |
| **Polish Bankruptcy** | 0.672 | 0.894 | 0.783 | 0.932 | 0.894 | **0.913** |

## 7.2 Comparative study of AUC

The best AUC results obtained in the respective datasets, generated by Vanilla CTGAN and modified CTGAN are presented in Table 7 and Table 8, respectively. We present the results for both the local training and meta-clustering phases. In addition to AUC, we also presented the number of centers.

In both cases, by employing Vanilla CTGAN and modified CTGAN, there is no significant change in AUC before and after the clustering phases. This indeed indicates that the employed meta-clustering is powerful enough in obtaining the generalized meta-centers, thereby reducing the row space dimension significantly. In terms of AUC, FedPNN achieved better AUC in all of the datasets using the synthetic dataset generated by modified CTGAN when compared to Vanilla CTGAN. Additionally, it is desired to obtain better AUC with the less possible number of cluster centers which indeed reflects the computational efficiency. Except in the Polish Bankruptcy dataset, the meta-clustering aggregation algorithm of FedPNN built on the synthetic dataset generated by modified CTGAN selected less number of centers in the negative class than that of Vanilla CTGAN. This demonstrates that employing modified CTGAN during synthetic dataset generation indeed preserves privacy and obtains higher AUC with less complexity when compared to Vanilla CTGAN. This is due to the fact that employing multiple distributions based on modified CTGAN captured the complex dependencies between the features better than Vanilla CTGAN.

**Table 7**: Comparative analysis on AUC after employing FedPNN

| Dataset | Type of the Node | Vanilla CTGAN | | | | Modified CTGAN | | | |
|---|---|---|---|---|---|---|---|---|---|
| | | Local Training | | After Meta-clustering | | Local Training | | After Meta-clustering | |
| | | AUC | No of Centers (-ve; +ve) | AUC | No of Centers (-ve; +ve) | AUC | No of Centers (-ve; +ve) | AUC | No of Centers (-ve; +ve) |
| Breast Cancer | $Client_1$ | 1 | (1;27) | 0.976 | (3; 17) | 0.989 | (2; 27) | **1** | **(2; 18)** |
| | $Client_2$ | 0.988 | (2; 28) | 0.927 | | 1 | (3; 27) | **1** | |
| | Server | - | - | 0.957 | | - | - | 0.989 | |
| Ova Omentum | $Client_1$ | 0.746 | (137; 16) | 0.735 | (276; 21) | 0.989 | (135; 1) | 0.989 | (267; 6) |
| | $Client_2$ | 0.798 | (139; 5) | 0.794 | | 0.973 | (133; 4) | 0.973 | |
| | Server | - | - | 0.738 | | - | - | 0.966 | |
| Credit card fraud | $Client_1$ | 0.933 | (11;148) | 0.911 | (13; 74) | 0.985 | (5; 213) | 0.947 | (4; 117) |
| | $Client_2$ | 0.94 | (17; 148) | 0.912 | | 0.987 | (4; 209) | 0.946 | |
| | Server | - | - | 0.914 | | - | - | 0.949 | |
| Polish Bankruptcy | $Client_1$ | 0.945 | (16; 37) | 0.947 | **(21; 64)** | 0.977 | (27; 67) | 0.979 | (36; 110) |
| | $Client_2$ | 0.897 | (11; 38) | 0.964 | | 0.965 | (28; 70) | 0.982 | |
| | Server | - | - | 0.853 | | - | - | 0.899 | |

\* -ve : Negative class ; +ve : Positive class ; - : No local training in Server

## 7.3 Sensitivity Analysis

In this section, we presented the sensitivity analysis conducted based on client $D_{thr}$ and server $D_{thr}$. Here, we reported both the change occurred in AUC and the number of centers. The sensitivity analysis was conducted in two different ways: (i) by fixing the best client Dthr and varying the server Dthr parameter with an interval of 0.02 is depicted in Fig. 7 to Fig. 10, Fig. 7 (a) denotes the change in AUC, while Fig. 7 (b) denotes the change in the number of centers for the Breast Cancer dataset. The same is followed for the rest of the datasets as shown in Fig. 8, Fig. 9 and Fig. 10. (ii) by fixing the best server Dthr and varying the client Dthr parameter with an interval of 0.02, is depicted in Fig. 11 through Fig. 14. It is important note that the sensitivity analysis is restricted to the synthetic dataset generated by modified CTGAN due to its better performance, as discussed in Section 7.2.

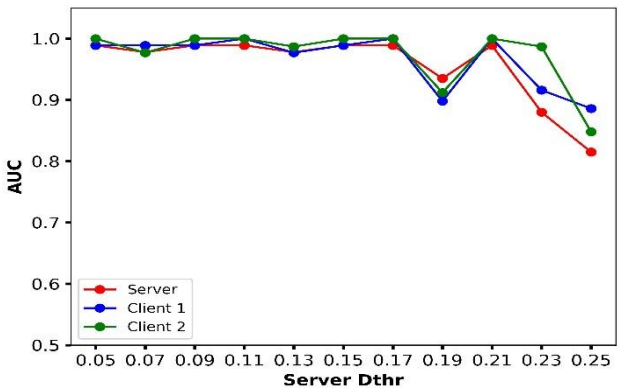
Fig. 7 (a) Change in AUC for Breast cancer by keeping client Dthr constant

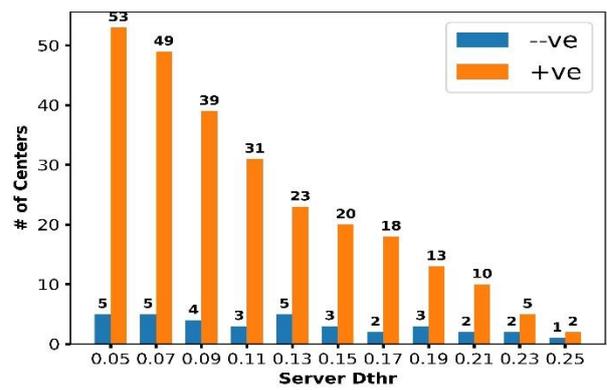
Fig. 7 (b) Change in number of centers for Breast cancer by keeping client Dthr constant

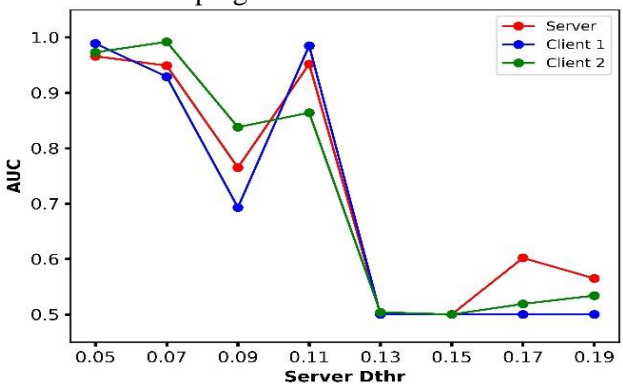
Fig. 8 (a) Change in AUC for OVA Omentum by keeping client Dthr constant

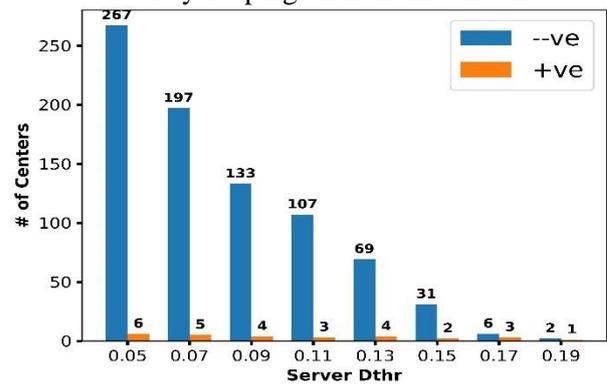
Fig. 8 (b) Change in the number of centers for OVA Omentum by keeping client Dthr constant

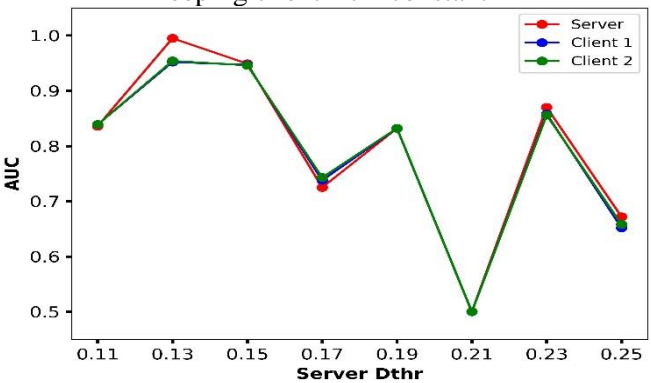
Fig. 9 (a) Change in AUC for Credit Card Fraud by keeping client Dthr constant

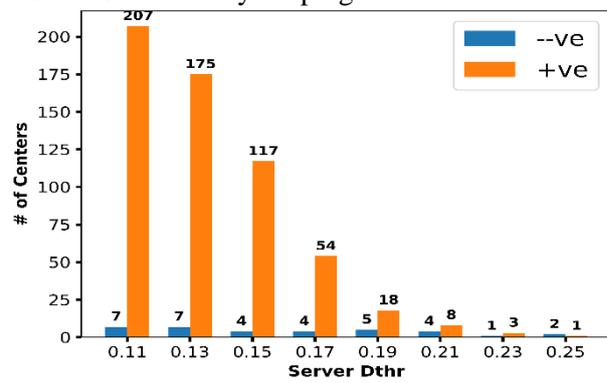
Fig. 9 (b) Change in the number of centers for Credit Card Fraud by keeping client Dthr constant

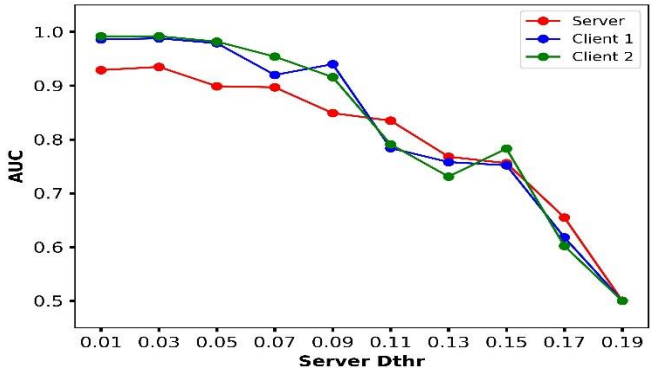 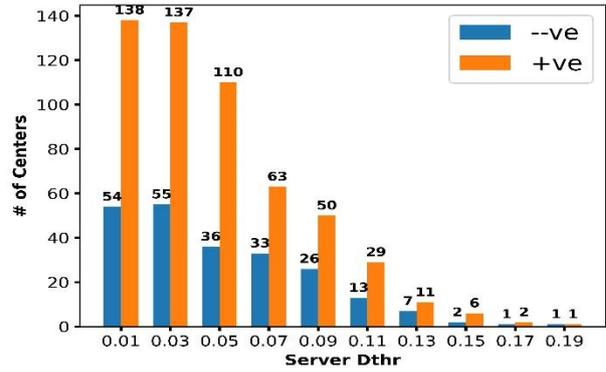

| Fig. 10 (a) Change in AUC for Polish Bankruptcy by keeping client Dthr constant | Fig. 10 (b) Change in the number of centers for Polish Bankruptcy by keeping client Dthr constant |
|---|---|

Firstly, we discussed the performance of FedPNN by choosing the best value for client Dthr and keeping it constant but with varied server Dthr parameters. For the Breast cancer dataset, the best server Dthr is 0.17 (refer to Table 3), it is observed that as there is an increase in server Dthr, there is a decrease in AUC. Although no specific trend is observed, the AUC constantly decreased after 0.21. However, there is no significant variation is observed while decreasing the server Dthr value i.e., in 0.15, 0.13, 0.11, etc. Similarly, for Credit card fraud (see Fig. 9) and Polish Bankruptcy datasets (see Fig. 10), overall, a decreasing trend in AUC is observed while increasing the server Dthr parameter. Further, in terms of change in the number of centers, the number of negative centers is constantly decreasing while increasing the server Dthr. However, no particular trend is observed for the number of positive class centers. Interestingly, except for the OVA Omentum dataset (see Fig. 8), the number of negative class centers is more than the number of positive class centers. As the number of centers is indeed responsible for the complexity of PNN, choosing an optimal number of centers that maintain good AUC is critical. For example, in the Breast cancer dataset (see Fig. 7), AUC obtained with 18 number of centers turned out to be almost similar to the AUC achieved when the number of centers are 53. Similarly, in the case of the OVA Omentum dataset, AUC achieved with 197 centers is slightly higher than with that of 267 centers, resulting in a significant decrease in the model complexity. Similar kind of observations, such as trends in AUC with respect to the number of centers are made for Credit card fraud and Polish Bankruptcy datasets.

Now, we will discuss the performance of FedPNN by keeping the best server Dthr value constant (refer to Table 3) and changing the client Dthr. Similar to the observations of the first analysis, here also, (i) there is a decrement in AUC when there is an increase in the client Dthr value in all datasets, and (ii) the number of centers is decreased as there is an increase in client Dthr. Further, in the credit card dataset (see Fig. 13), the AUC achieved with 965 centers is lower than that of 470 centers. This concludes the fact the AUC is susceptible to change in server Dthr and client Dthr. Hence, with the above analysis, we conclude that appropriate values need to be chosen based on considering both AUC and the number of centers.

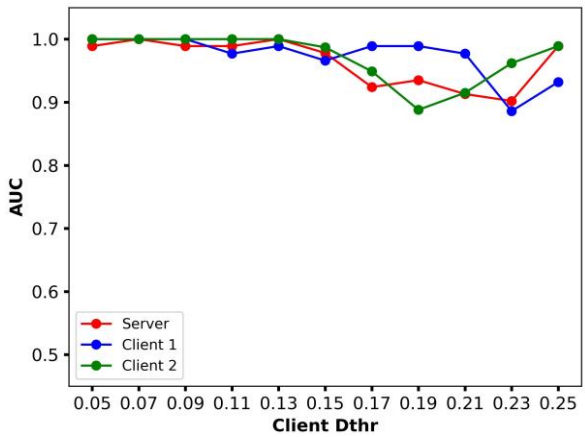
Fig. 11 (a) Change in AUC for Breast cancer by keeping server Dthr constant

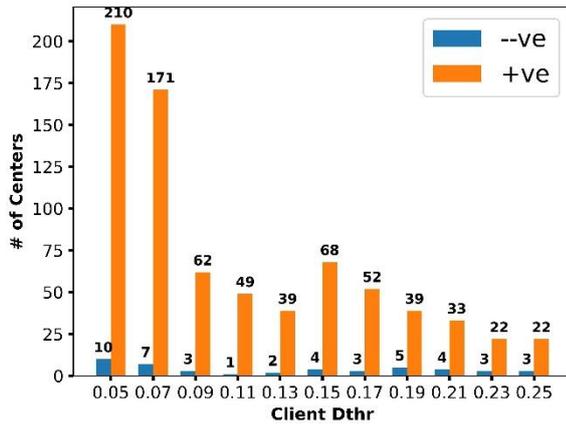
Fig. 11 (b) Change in the number of centers for Breast cancer by keeping server Dthr constant

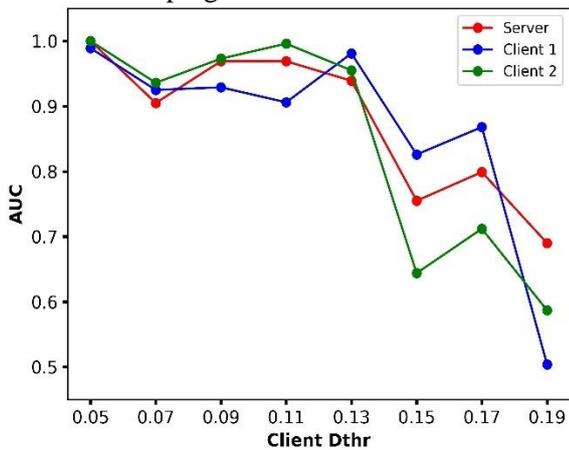
Fig. 12 (a) Change in AUC for OVA Omentum by keeping server Dthr constant

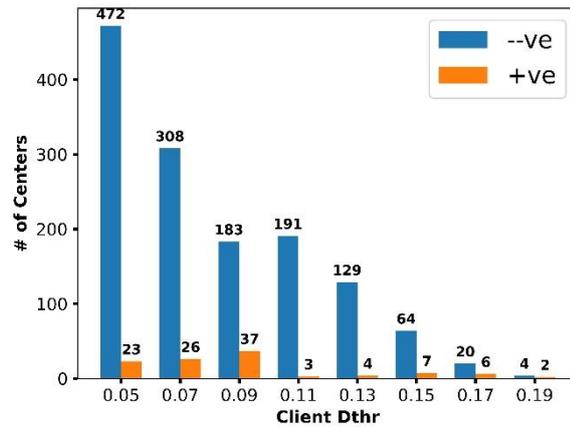
Fig. 12 (b) Change in the number of centers for Breast cancer by keeping server Dthr constant

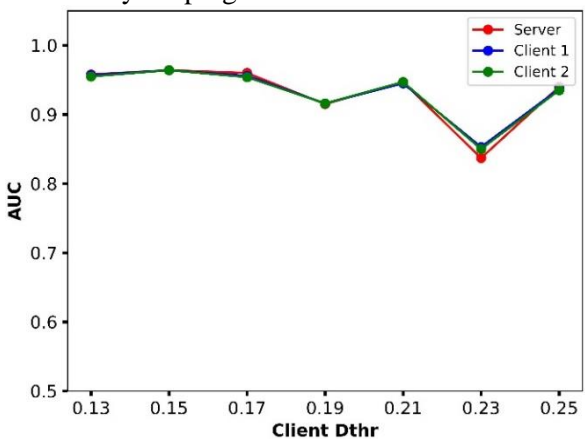
Fig. 13 (a) Change in AUC for Credit Card Fraud by keeping server Dthr constant

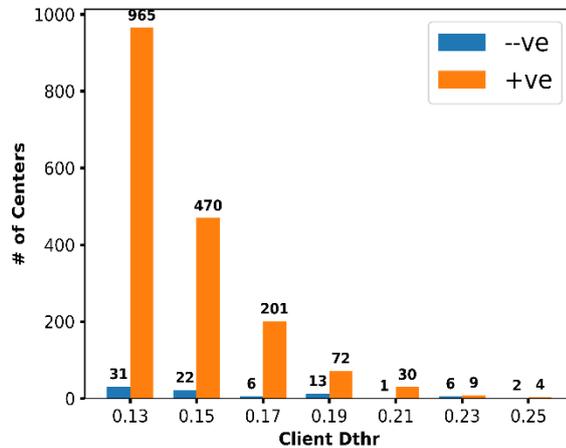
Fig. 13 (b) Change in the number of centers for Credit Card Fraud by keeping server Dthr constant

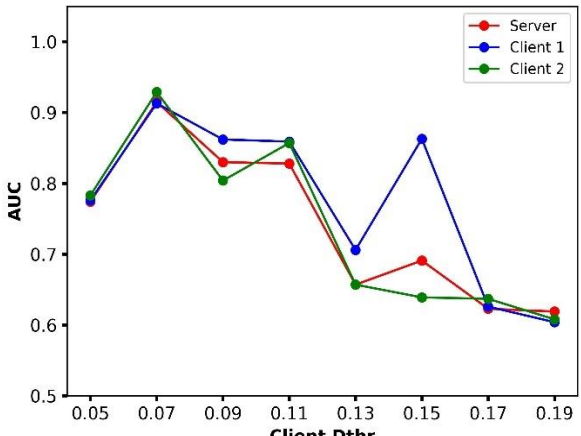 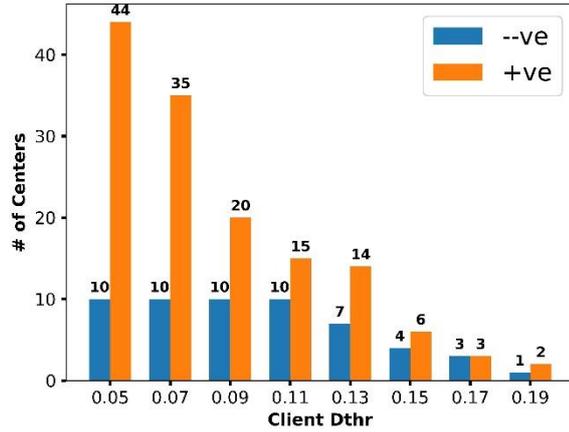

| Fig. 14 (a) Change in AUC for Polish Bankruptcy by keeping server Dthr constant | Fig. 14 (b) Change in AUC for Polish Bankruptcy by keeping server Dthr constant |

## 8. Conclusions and Future Directions

In the current study, we proposed a two-stage one-shot federated learning approach toward the objective of privacy protection and overcoming a few challenges of FL. Here, during the first stage, we employed a modified CT-GAN to generate the synthetic dataset, which is later evaluated by the KSComplement and CStest scores. The thus generated synthetic dataset is distributed among the nodes. Thereafter, FedPNN is employed and integrated with modified ECM, and at the server, the meta-aggregation algorithm is invoked during the communication round. The results demonstrate that the proposed modified CTGAN obtained better CStest scores, 3 out of 4 times but outperformed by Vanilla CTGAN in terms of KSComplement score in the majority of the cases. Further, the proposed meta-clustering aggregation algorithm is able to generalize well as there is only a slight fluctuation of AUC before and after its invocation but decreased the complexity by selecting less number of optimal centers in 3 out of 4 datasets. Additionally, the sensitivity analysis demonstrated to us how the number of centers is influencing AUC and trends in AUC and the number of centers.

The following are potential future works: (i) Employing FedCTGAN in the architecture to improve better privacy, but doing so increases the complexity and hence it should be taken care of, (ii) Handling highly imbalanced datasets, (iii) Developing regression based models along the similar lines, (iv) Designing and employing homomorphic encryption based FedPNN in the above settings to increase the level of privacy protection. Additionally, this is a general purpose methodology which can be employed in the following use cases in banking, finance and insurance sector which are as follows: (a) anti-money laundering, (b) detecting multiple asset creation, (c) identifying frauds in insurance claims, (d) identifying impact of geographical differences on diseases, (e) fraud detection in intra-day stock trading.